\documentclass[fancyhdr]{article}

\usepackage{authblk}
\usepackage[utf8]{inputenc}
\usepackage[T1]{fontenc}
\usepackage{pslatex}
\usepackage[english]{babel}
\usepackage{fancyhdr}
\usepackage{hyperref}
\usepackage{amsmath,amssymb,amsfonts,amsthm}
\usepackage{graphicx}
\usepackage{amsmath}
\usepackage{amsthm}
\usepackage{amsfonts}
\usepackage{mathtools}
\usepackage{amssymb}
\usepackage{algorithm}
\usepackage{algorithmic}
\usepackage{microtype}
\usepackage{rotating}
\usepackage{todonotes}
\usepackage{tikz}
\usepackage{tikz-qtree}
\usepackage{xparse}
\usepackage{appendix}
\usepackage{url}
\usepackage{tabularx}
\usepackage{paralist}
\usepackage{booktabs}
\usepackage{multirow}
\usepackage{subcaption}
\usepackage{cancel}
\usetikzlibrary{calc,shapes.callouts,shapes.arrows}
\usetikzlibrary{automata}
\usetikzlibrary{positioning}
\usetikzlibrary{decorations.pathreplacing}
\usetikzlibrary{decorations.pathmorphing}
\usetikzlibrary{arrows}
\usetikzlibrary{shapes}

\newcommand{\F}{\mathbb{F}}
\newcommand{\Z}{\mathbb{Z}}

\newcommand{\wt}{\mathrm{wt}}

\newcommand{\Nega}{\mathcal{N}}

\newcommand{\sigmaone}{\sigma_1}
\newcommand{\sigmatwo}{\sigma_2}

\newtheorem*{problem*}{Problem}

\providecommand{\keywords}[1]{\textbf{\textit{Keywords }} #1}

\begin{document}

\title{NegaBent, No Regrets: Evolving Spectrally Flat Boolean Functions}

\author[1,2]{Claude Carlet}
\author[3]{Marko \DH urasevic}
\author[2]{Ermes Franch}
\author[3]{Domagoj Jakobovic}
\author[4]{Luca Mariot}
\author[3,5]{Stjepan Picek}

\affil[1 ]{{\normalsize University of Paris 8, Saint-Denis, France}}

\affil[2 ]{{\normalsize University of Bergen, Bergen, Norway}

    {\small \texttt{claude.carlet@gmail.com, ermes.franch@uib.no}}}

\affil[3 ]{{\normalsize Faculty of Electrical Engineering and Computing, University of Zagreb, Unska 3, Zagreb, Croatia} \\

{\small \texttt{marko.durasevic@fer.hr, domagoj.jakobovic@fer.hr}}}

\affil[4 ]{{\normalsize Semantics, Cybersecurity and Services Group, University of Twente, 7522 NB Enschede, The Netherlands} \\
	
	{\small \texttt{l.mariot@utwente.nl}}}

\affil[5 ]{{\normalsize Digital Security Group, Radboud University, Postbus 9010, 6500 GL Nijmegen, The Netherlands} \\
	
	{\small \texttt{stjepan@computer.org}}}
	
\maketitle

\begin{abstract}
Negabent Boolean functions are defined by having a flat magnitude spectrum under the nega-Hadamard transform.
They exist in both even and odd dimensions, and the subclass of functions that are simultaneously bent and negabent
(bent-negabent) has attracted interest due to the combined optimal periodic and negaperiodic spectral properties.
In this work, we investigate how evolutionary algorithms can be used to evolve (bent-)negabent Boolean functions. 
Our experimental results indicate that evolutionary algorithms, especially genetic programming, are a suitable approach for evolving negabent Boolean functions, and we successfully evolve such functions in all dimensions we consider.
\end{abstract}


\section{Introduction}
\label{sec:introduction}
Boolean functions have an important role in diverse domains, like cryptography~\cite{carlet_2021}, coding theory~\cite{KERDOCK1972182,MacWilliams-Sloane}, sequence design~\cite{1056589}, and combinatorics~\cite{Rothaus}.
Naturally, to be useful, Boolean functions need to fulfill a number of properties, where having high nonlinearity is among the most common ones~\cite{carlet_2021}.
Since the search space of all Boolean functions is large already for moderate input size $n$ ($2^{2^n}$), one needs to approach constructing Boolean functions with good properties in a structured way. Today, two common approaches are to use algebraic constructions and metaheuristics/heuristics.

In even dimensions, Boolean functions can be maximally nonlinear, and such functions are called bent. 
Negabent functions are an analogous notion for the nega-Hadamard transform, motivated in part by local unitary transforms in quantum-information settings and also studied for their cryptographic/coding relevance~\cite{6457455,10.1007/978-3-642-15874-2_31}.
A Boolean function that is both bent and negabent is called bent-negabent and combines optimality under two distinct spectral views.
Negabent and bent-negabent functions have been studied for their spectral/transform properties, connections to negaperiodic correlation, and appearance in generalized bent frameworks motivated by quantum/stabilizer analysis~\cite{riera2005generalisedbentcriteriaboolean}.
In cryptographic design, they are relevant as structured candidates for high nonlinearity and good correlation behavior, and as
building blocks in classes where multiple spectral constraints are desired (e.g., combining different transform flatness notions)~\cite{10.1007/978-3-540-85912-3_34}.

\textbf{Related Work.} 
Using metaheuristics to evolve Boolean functions is a well-accepted and powerful approach. Moreover, obtaining high nonlinearity represents the most usual task~\cite{Djurasevic2023}.
Bent Boolean functions are a common target for metaheuristics because they have practical relevance and a well-defined maximal possible nonlinearity, allowing one to clearly assess how well the optimization process works. 
For instance, Fuller et al. used evolutionary algorithms to design bent Boolean functions more than 20 years ago~\cite{FullerDM03}.
However, current results indicate that evolving a bent Boolean function is not necessarily difficult, and the main problem is the computational complexity of evaluating the nonlinearity property. To alleviate that problem, Hrbacek and Dvorak used Cartesian Genetic Programming to speed up the evolution process and evolve large bent Boolean functions~\cite{10.1007/978-3-319-10762-2_41}. On the other hand, Picek and Jakobovic proposed an evolutionary algorithm approach to evolve secondary constructions of bent Boolean functions~\cite{10.1145/2908812.2908915}.
Husa and Dobai proposed using linear genetic programming to evolve bent Boolean functions, citing better performance than genetic programming for larger sizes~\cite{10.1145/3067695.3084220}.
Besides considering general bent Boolean functions, several works consider bentness in specific settings. For instance, Picek et al. evolved quaternary bent Boolean functions~\cite{picek18a}, while Mariot et al. considered hyper-bent Boolean functions~\cite{hyperbent}. Carlet et al. considered evolving rotation symmetric self-dual bent functions~\cite{10.1007/978-3-031-70085-9_27} and homogenous bent functions~\cite{10.1145/3712255.3726779}. 
Interestingly, evolving highly nonlinear Boolean functions in odd dimensions is a much less considered topic, and we point interested readers to a recent work indicating evolutionary algorithms as performing well (and on level of specialized heuristics)~\cite{Carlet2025}.

\textbf{Main Contributions.} 
In this work, we consider the problem of evolving negabent Boolean functions. More precisely, for even dimensions, we evolve bent-negabent Boolean functions, and for odd dimensions, we evolve highly nonlinear negabent Boolean functions. 
Unlike bentness (which excludes affine functions except in trivial cases), all affine Boolean functions are negabent~\cite{10.1007/978-3-642-10628-6_9}.
This has consequences for evolutionary search since negabentness alone does not guarantee high nonlinearity, as the search could simply produce affine functions (which are linear).
In our experiments, we employ two solution encodings and one fitness function and show that this problem is well within the reach of evolutionary algorithms. Indeed, we find highly (maximally) nonlinear negabent Boolean functions for each tested odd and even dimension, respectively.

\section{Boolean Functions - Representations and Properties}
\label{sec:background}

We denote by $\mathbb F_2=\{0,1\}$ the finite field with two elements, equipped with XOR and logical AND, respectively, as the sum and multiplication operations. The $n$-dimensional vector space over $\mathbb F_2$ is denoted by $\mathbb F_2^n$, consisting of all $2^n$ binary vectors of length $n$. Given $a, b \in \mathbb F_2^n$, their inner product equals $a\cdot b = \bigoplus_{i=1}^{n} a_{i}b_{i}$ in $\mathbb F_{2}^n$. A Boolean function of $n$ variables is a mapping $f: \F_2^n \to \F_2$. 

A Boolean function $f$ can be uniquely represented by its truth table. The truth table of a Boolean function $f$ is the list of pairs $(x, f(x))$ of input vectors $x \in \F_2^n$ and function outputs $f(x) \in \F_2$. 

A Boolean function $f$ can also be uniquely represented by the Walsh-Hadamard transform $W_{f}: \F_2^n \to \Z$. The Walsh-Hadamard transform measures the correlation between $f$ and the linear functions $a\cdot x$, for all $a \in \mathbb F_2^n$:
\begin{equation}
W_{f} (a) = \sum\limits_{x \in \mathbb{F}_{2}^{n}} (-1)^{f(x) \oplus a\cdot x},
\end{equation}
where the sum is calculated in ${\mathbb Z}$. To calculate the Walsh-Hadamard spectrum, it is common to use an efficient butterfly algorithm with complexity $O(n2^n)$.

The minimum Hamming distance between a Boolean function $f$ and all affine functions is the nonlinearity of $f$, calculated from the Walsh-Hadamard spectrum as~\cite{carlet_2021}:
\begin{equation}
\label{eq:nonlinearity}
nl_{f} = 2^{n - 1} - \frac{1}{2}\max_{a \in \mathbb{F}_{2}^{n}} \left \{ |W_{f}(a)| \right \}.
\end{equation}

For every $n$-variable Boolean function, the function $f$ satisfies the covering radius bound:
\begin{equation}
\label{eq_boolean_covering}
    nl_{f} \leq 2^{n-1}-2^{\frac n 2 - 1}.
\end{equation}
Notice that Eq.~\eqref{eq_boolean_covering} cannot be tight when $n$ is odd. In other words, maximal nonlinearity is possible for even $n$ only (such functions are called bent Boolean functions). 

The maximal possible nonlinearity for odd-sized Boolean functions lies between~\cite{carlet_2021}:
$2^{n-1}-2^{\frac{n-1}{2}}$ and $2\lfloor2^{n-2}-2^{\frac n 2-2}\rfloor$, 
which will also represent the nonlinearity range for negabent functions in an odd number of variables.

Let $i^2=-1$. We define the nega-Hadamard transform by
\begin{equation}
\label{eq:nega-def}
\Nega_f(a)=\sum_{x\in\F_2^n} (-1)^{f(x)\oplus a \cdot x} i^{\wt(x)},\qquad a\in\F_2^n,
\end{equation}
where $\wt(x)$ is the Hamming weight~\cite{6457455}.
A Boolean function $f$ is negabent if and only if $|\Nega_f(a)|=2^{\frac n 2}$ for all $a\in\F_2^n$~\cite{6457455,10.1007/978-3-642-15874-2_31}.
In an even dimension $n$, a function is bent-negabent if and only if it is bent and negabent. 

When $n$ is even, testing negabentness reduces to a standard bent test~\cite{6457455,10.1007/978-3-540-77404-4_2}.

Then $f$ is negabent if and only if $f\oplus\sigmatwo$ is bent.
Here, for $x=(x_1,\dots,x_n)\in\F_2^n$, define
\[
\sigmatwo(x)=\bigoplus_{1\le i<j\le n} x_ix_j.
\]
Moreover, if $f$ is a bent-negabent function, then $f\oplus \sigmatwo$ is also bent-negabent~\cite{6457455}.

For odd $n$, the function $f(x)$ is negabent if 
$f(x)\oplus \sigmatwo(x)\oplus \sigmaone(x)y$ is bent in $(n+1)$ variables, where $x \in \mathbb F^n_2$ and $y \in \mathbb F_2$~\cite{6457455}.

Here, for $x=(x_1,\dots,x_n)\in\F_2^n$:
\[
\sigmaone(x)=\bigoplus_{j=1}^n x_j.
\]

Note that all affine functions\footnote{An affine function has the algebraic degree at most 1~\cite{carlet_2021}.} (both with an even and an odd number of variables) are negabent~\cite{10.1007/978-3-540-77404-4_2}. For even $n \ge 2$, bent functions have maximum nonlinearity and are not affine; as such, nontrivial bent-negabent functions are non-affine.

\section{Experimental Setup}
\label{sec:setup}

\subsection{Bitstring Encoding}
A straightforward option for encoding a Boolean function is the bitstring, which represents the function's truth table (TT). For a Boolean function with $n$ inputs, the truth table is coded as a bitstring with a length of $2^n$.
For each evaluation, the truth table is taken and tested for the bent-negabent property, depending on whether the number of variables is odd or even, as described in the previous section.

\subsection{Symbolic Encoding}
The second approach in our experiments uses tree-based genetic programming (GP) to represent a Boolean function in its symbolic form.
In this case, a candidate solution is represented by a tree whose leaf nodes correspond to the input variables $x_1,\cdots, x_n \in \F_2$. The internal nodes are Boolean operators that combine the inputs received from their children and forward their output to the respective parent nodes.
The output of the root node is the output value of the Boolean function. The corresponding truth table of $f: \F_2^n \to \F_2$ is determined by evaluating the tree over all possible $2^n$ assignments of the input variables at the leaf nodes. 
Each individual is evaluated according to the truth table it generates, in the same way as for the bitstring representation.

\subsection{Fitness Function}
Since we optimize nonlinearity, in solution evaluation for an even number of variables, we need to calculate both the nonlinearity of the original function $nl_f$, as well as the nonlinearity of $f\oplus\sigmatwo$, denoted as $nl_{f\oplus\sigmatwo}$. 
Since both of these need to be at the upper bound defined in Eq.~\eqref{eq_boolean_covering}, a simple approach would be to define the fitness function as the sum of these values, $nl_f + nl_{f\oplus\sigmatwo}$.
However, in our experiments, we consider the whole Walsh-Hadamard spectrum and not only its extreme value (see Eq.~\eqref{eq:nonlinearity}).
To provide some form of gradient information, we include the number of occurrences of the maximal absolute value in the spectrum of both functions, denoted as $\#max\_values_f$ and $\#max\_values_{f\oplus\sigmatwo}$ correspondingly.
Since higher nonlinearity corresponds to a \textit{lower} maximal absolute value, we aim for as few occurrences of the maximal value as possible to make it easier for the algorithm to reach the next nonlinearity value.
Therefore, the fitness function for even $n$ is defined as:
\begin{equation}
\label{eq:fitness}
fitness = nl_{f} + \frac{2^n - \#max\_values_f}{2^n} + nl_{f\oplus\sigmatwo} + \frac{2^n - \#max\_values_{f\oplus\sigmatwo}}{2^n}
\end{equation}
Note that the terms in fractions never reach the value of $1$ since, in that case, we effectively reach the next nonlinearity level.
The optimal value of this fitness function, obtained if both functions are bent, is equal to twice the upper bound~\eqref{eq_boolean_covering}. 

For an odd number of variables $n$, we first construct the truth table of $f(x)\oplus \sigmatwo(x)\oplus \sigmaone(x)y$ as described previously, and just maximize the nonlinearity of the resulting function with an even number of variables $n+1$, also considering the whole Walsh-Hadamard spectrum:
\begin{equation}
\label{eq:fitness_odd}
fitness = nl_{f(x)\oplus \sigmatwo(x)\oplus \sigmaone(x)y} + \frac{2^n - \#max\_values}{2^n} 
\end{equation}
In this case, the optimal fitness value is equal to the covering radius, corresponding to a bent function in $n+1$ variables.

\subsection{Algorithm Parameters}

\textbf{Bitstring Encoding.}
The corresponding variation operators we use are the simple bit mutation and the shuffle mutation.
The simple bit mutation inverts a randomly selected bit. The shuffle mutation shuffles the bits within a randomly selected substring.
For the crossover operators, we use the one-point crossover and uniform crossover. 
The one-point crossover combines a new solution from the first part of one parent and the second part of the other parent with a randomly selected breakpoint.
The uniform crossover randomly selects one bit from both parents at each position in the child bitstring that is copied.
Each time the evolutionary algorithm invokes a crossover or mutation operation, one of the previously described operators is randomly selected.

\textbf{Symbolic Encoding.}
In our experiments, we use the following function set that takes two arguments: OR, XOR, and AND. Next, we use the function NOT that takes a single argument.
Finally, we use the function IF that takes three arguments and returns the second one if the first one evaluates to true, and the third one otherwise. This function set is common when dealing with the evolution of Boolean functions with cryptographic properties~\cite{Djurasevic2023,10.1007/978-3-031-70085-9_27}.

The genetic operators used in our experiments with tree-based GP are simple tree crossover, uniform crossover, size fair, one-point, and context preserving crossover~\cite{Poli2008} (selected at random), and subtree mutation.

We employ the same evolutionary algorithm for both bitstring and symbolic encoding: a steady-state selection with a 3-tournament elimination operator (denoted SST). 
In each iteration of the algorithm, three individuals are chosen at random from the population for the tournament, and the worst one in terms of fitness value is eliminated. 
The two remaining individuals in the tournament are used with the crossover operator to generate a new child individual, which then undergoes mutation with individual mutation probability $p_{mut} = 0.5$. Finally, the mutated child replaces the eliminated individual in the population.
The population size in all experiments was 500 individuals.

\section{Experimental Results}
\label{sec:results}

\begin{figure*}
    \centering
    \includegraphics[width=\linewidth]{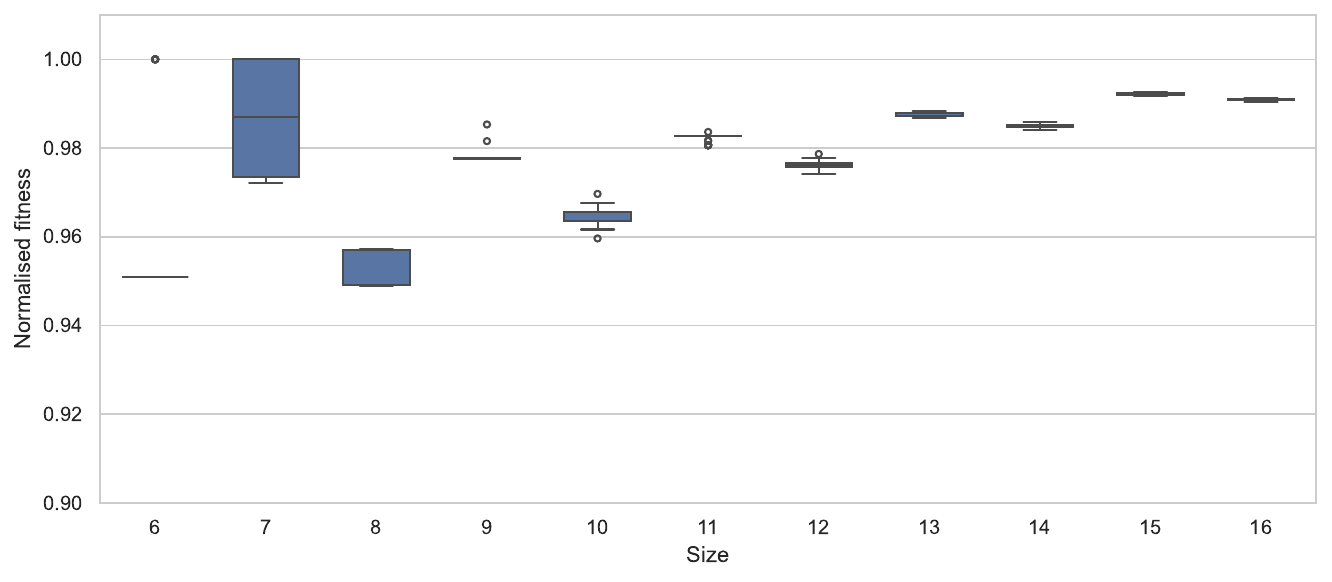}
    \caption{Optimization results for bitstring encoding (truth table, TT)}
    \label{fig:tt}
\end{figure*}

\begin{figure*}
    \centering
    \includegraphics[width=\linewidth]{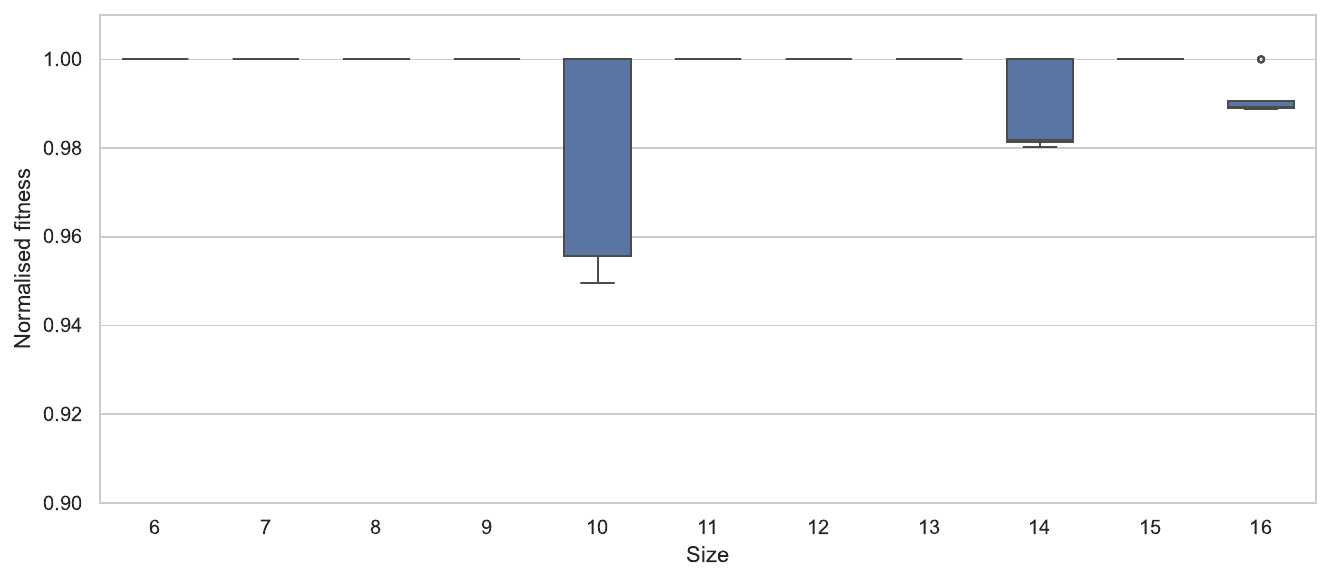}
    \caption{Optimization results for symbolic encoding (tree, GP)}
    \label{fig:gp}
\end{figure*}

The experiments were conducted with the parameter settings described above; each experiment
was executed 30 times, and the termination condition was set at $10^6$ evaluations for all configurations.
We present the results in the form of a separate boxplot for each of the encodings (TT and GP) for all Boolean function sizes from 6 to 16.
Since the optimal objective values are known, we normalize the plot so that the optimal value equals 1 in all sizes and apply a suitable scale to enhance visibility.
The aggregated results are shown in Figures~\ref{fig:tt} and~\ref{fig:gp}.

The TT encoding found bent-negabent functions only in sizes 6 and 7. On the other hand, the GP encoding found such functions for every number of variables in at least one of 30 runs.
Except for sizes 10, 14, and 16, a bent-negabent function was obtained in every run.
Clearly, this indicates that the fitness functions were well defined, which is expected as we modeled them based on related work. Moreover, since the related work demonstrated that evolving bent Boolean functions is well within the reach of diverse evolutionary algorithms, it is not surprising that we had no difficulty in evolving bent-negabent functions.
For odd dimensions, the results are more interesting as the related work also provides less references for maximally nonlinear functions. What is more, the fitness function for odd sizes is more complicated, potentially disrupting the connection between what we evolve (in $n$ variables) and what is evaluated (in $n+1$ variables). The excellent behavior for odd dimensions suggests we find strong structure during the evolution that remains present for evaluation.

\section{Conclusions and Future Work}
\label{sec:conclusions}

This paper considers the problem of evolving bent-negabent Boolean functions for even sizes and highly nonlinear negabent functions for odd sizes. 
To our knowledge, this is the first time that the evolution of negabent function is considered. 
One challenge (besides computational complexity of evaluating the solution for higher dimensions) that could be expected when evolving negabent functions is that we use efficient tests that check bentness of transformed functions so to avoid working with complex values. As such, the transformed function must maintain the information that drives the evolution process.
Our results indicate this to be difficult problem for the truth table encoding, but an easy problem for the symbolic encoding. Indeed, with GP, we find highly (maximally) nonlinear negabent functions for all evaluated Boolean function sizes. 

In future work, it would be interesting to evolve negabent functions and evaluate their nonlinearity through the nega-Hadamard transform. Moreover, a natural direction would be to consider additional properties like belancedness and algebraic degree.

\bibliographystyle{abbrv}
\bibliography{bibliography}

\begin{thebibliography}{10}

\bibitem{carlet_2021}
C.~Carlet.
\newblock {\em Boolean Functions for Cryptography and Coding Theory}.
\newblock Cambridge University Press, Cambridge, 2021.

\bibitem{10.1145/3712255.3726779}
C.~Carlet, M.~Durasevic, D.~Jakobovic, L.~Mariot, and S.~Picek.
\newblock Degree is important: On evolving homogeneous boolean functions.
\newblock In {\em Proceedings of the Genetic and Evolutionary Computation
  Conference Companion}, GECCO '25 Companion, page 795–798, New York, NY,
  USA, 2025. Association for Computing Machinery.

\bibitem{10.1007/978-3-031-70085-9_27}
C.~Carlet, M.~Durasevic, D.~Jakobovic, and S.~Picek.
\newblock Discovering rotation symmetric self-dual bent functions with
  evolutionary algorithms.
\newblock In M.~Affenzeller, S.~M. Winkler, A.~V. Kononova, H.~Trautmann,
  T.~Tu{\v{s}}ar, P.~Machado, and T.~B{\"a}ck, editors, {\em Parallel Problem
  Solving from Nature -- PPSN XVIII}, pages 429--445, Cham, 2024. Springer
  Nature Switzerland.

\bibitem{Carlet2025}
C.~Carlet, M.~Durasevic, D.~Jakobovic, S.~Picek, and L.~Mariot.
\newblock A systematic study on the design of odd-sized highly nonlinear
  boolean functions via evolutionary algorithms.
\newblock {\em Genetic Programming and Evolvable Machines}, 27(1), Dec. 2025.

\bibitem{Djurasevic2023}
M.~Djurasevic, D.~Jakobovic, L.~Mariot, and S.~Picek.
\newblock A survey of metaheuristic algorithms for the design of cryptographic
  {B}oolean functions.
\newblock {\em Cryptography and Communications}, 15(6):1171--1197, July 2023.

\bibitem{FullerDM03}
J.~Fuller, E.~Dawson, and W.~Millan.
\newblock Evolutionary generation of bent functions for cryptography.
\newblock In {\em Proceedings of the {IEEE} Congress on Evolutionary
  Computation, {CEC} 2003, Canberra, Australia, December 8-12, 2003}, pages
  1655--1661. {IEEE}, 2003.

\bibitem{10.1007/978-3-319-10762-2_41}
R.~Hrbacek and V.~Dvorak.
\newblock Bent function synthesis by means of cartesian genetic programming.
\newblock In T.~Bartz-Beielstein, J.~Branke, B.~Filipi{\v{c}}, and J.~Smith,
  editors, {\em Parallel Problem Solving from Nature -- PPSN XIII}, pages
  414--423, Cham, 2014. Springer International Publishing.

\bibitem{10.1145/3067695.3084220}
J.~Husa and R.~Dobai.
\newblock Designing bent {B}oolean functions with parallelized linear genetic
  programming.
\newblock In {\em Proceedings of the Genetic and Evolutionary Computation
  Conference Companion}, GECCO '17, page 1825–1832, New York, NY, USA, 2017.
  Association for Computing Machinery.

\bibitem{KERDOCK1972182}
A.~Kerdock.
\newblock A class of low-rate nonlinear binary codes.
\newblock {\em Information and Control}, 20(2):182 -- 187, 1972.

\bibitem{MacWilliams-Sloane}
F.~J. MacWilliams and N.~J.~A. Sloane.
\newblock {\em The Theory of Error-Correcting Codes}.
\newblock Elsevier, Amsterdam, North Holland, 1977.
\newblock {ISBN: 978-0-444-85193-2}.

\bibitem{hyperbent}
L.~Mariot, D.~Jakobovic, A.~Leporati, and S.~Picek.
\newblock Hyper-bent boolean functions and evolutionary algorithms.
\newblock In L.~Sekanina, T.~Hu, N.~Louren{\c{c}}o, H.~Richter, and
  P.~Garc{\'i}a-S{\'a}nchez, editors, {\em Genetic Programming}, pages
  262--277, Cham, 2019. Springer International Publishing.

\bibitem{1056589}
J.~{Olsen}, R.~{Scholtz}, and L.~{Welch}.
\newblock Bent-function sequences.
\newblock {\em IEEE Transactions on Information Theory}, 28(6):858--864,
  November 1982.

\bibitem{10.1007/978-3-540-77404-4_2}
M.~G. Parker and A.~Pott.
\newblock On boolean functions which are bent and negabent.
\newblock In S.~W. Golomb, G.~Gong, T.~Helleseth, and H.-Y. Song, editors, {\em
  Sequences, Subsequences, and Consequences}, pages 9--23, Berlin, Heidelberg,
  2007. Springer Berlin Heidelberg.

\bibitem{10.1145/2908812.2908915}
S.~Picek and D.~Jakobovic.
\newblock Evolving algebraic constructions for designing bent {B}oolean
  functions.
\newblock In {\em Proceedings of the Genetic and Evolutionary Computation
  Conference 2016}, GECCO ’16, page 781–788, New York, NY, USA, 2016.
  Association for Computing Machinery.

\bibitem{picek18a}
S.~Picek, K.~Knezevic, L.~Mariot, D.~Jakobovic, and A.~Leporati.
\newblock Evolving bent quaternary functions.
\newblock In {\em 2018 {IEEE} Congress on Evolutionary Computation, {CEC} 2018,
  Rio de Janeiro, Brazil, July 8-13, 2018}, pages 1--8. {IEEE}, 2018.

\bibitem{Poli2008}
R.~Poli, W.~B. Langdon, and N.~F. McPhee.
\newblock {\em A Field Guide to Genetic Programming}.
\newblock Lulu Enterprises, UK Ltd, United Kingdom, 2008.

\bibitem{riera2005generalisedbentcriteriaboolean}
C.~Riera and M.~G. Parker.
\newblock Generalised bent criteria for boolean functions (i), 2005.

\bibitem{Rothaus}
O.~Rothaus.
\newblock On “bent” functions.
\newblock {\em Journal of Combinatorial Theory, Series A}, 20(3):300 -- 305,
  1976.

\bibitem{10.1007/978-3-642-10628-6_9}
S.~Sarkar.
\newblock On the symmetric negabent boolean functions.
\newblock In B.~Roy and N.~Sendrier, editors, {\em Progress in Cryptology -
  INDOCRYPT 2009}, pages 136--143, Berlin, Heidelberg, 2009. Springer Berlin
  Heidelberg.

\bibitem{10.1007/978-3-540-85912-3_34}
K.-U. Schmidt, M.~G. Parker, and A.~Pott.
\newblock Negabent functions in the maiorana--mcfarland class.
\newblock In S.~W. Golomb, M.~G. Parker, A.~Pott, and A.~Winterhof, editors,
  {\em Sequences and Their Applications - SETA 2008}, pages 390--402, Berlin,
  Heidelberg, 2008. Springer Berlin Heidelberg.

\bibitem{10.1007/978-3-642-15874-2_31}
P.~St{\u{a}}nic{\u{a}}, S.~Gangopadhyay, A.~Chaturvedi, A.~K. Gangopadhyay, and
  S.~Maitra.
\newblock Nega--hadamard transform, bent and negabent functions.
\newblock In C.~Carlet and A.~Pott, editors, {\em Sequences and Their
  Applications -- SETA 2010}, pages 359--372, Berlin, Heidelberg, 2010.
  Springer Berlin Heidelberg.

\bibitem{6457455}
W.~Su, A.~Pott, and X.~Tang.
\newblock Characterization of negabent functions and construction of
  bent-negabent functions with maximum algebraic degree.
\newblock {\em IEEE Transactions on Information Theory}, 59(6):3387--3395,
  2013.

\end{thebibliography}

\end{document}